\documentclass{acm_proc_article-sp}

\usepackage{color}

\newcommand{\tdl}[0]{$\mathcal{DL}$-LTL}

\usepackage[amsmath,thref,thmmarks]{ntheorem}
\usepackage{mdwlist}
\usepackage{arydshln}
\usepackage{multirow}

\newtheorem{definition}{Definition}

\usepackage{algorithm}
\usepackage{algorithmic}

\begin{document}

\title{Enabling Semantic Analysis of User Browsing Patterns in the Web of Data}

%
%
%
%
%

\numberofauthors{3} 
%
\author{
%
%
\alignauthor
Julia Hoxha\\
       \affaddr{Institute of Applied Informatics and Formal Methods}\\
       \affaddr{Karlsruhe Institute of Technology}\\
       \affaddr{D-76128 Karlsruhe, Germany}\\
       \email{julia.hoxha@kit.edu}
\alignauthor
Martin Junghans\\
       \affaddr{Institute of Applied Informatics and Formal Methods}\\
       \affaddr{Karlsruhe Institute of Technology}\\
       \affaddr{D-76128 Karlsruhe, Germany}\\
       \email{martin.junghans@kit.edu}
\alignauthor 
Sudhir Agarwal\\
       \affaddr{Institute of Applied Informatics and Formal Methods}\\
       \affaddr{Karlsruhe Institute of Technology}\\
       \affaddr{D-76128 Karlsruhe, Germany}\\
       \email{sudhir.agarwal@kit.edu}
}
\date{14 February 2012}

\maketitle
\begin{abstract}
A useful step towards better interpretation and analysis of the usage patterns is to formalize the semantics of the resources that users are accessing in the Web. We focus on this problem and present an approach for the semantic formalization of usage logs, which lays the basis for effective techniques of querying expressive usage patterns. We also present a query answering approach, which is useful to find in the logs expressive patterns of usage behavior via formulation of semantic and temporal-based constraints.

We have processed over 30 thousand user browsing sessions extracted from usage logs of DBPedia and Semantic Web Dog Food. The logs are semantically formalized using respective domain ontologies and RDF representations of the Web resources being accessed. 
We show the effectiveness of our approach through experimental results, providing in this way an exploratory analysis of the way users browse the Web of Data.
\end{abstract}



\keywords{web browsing behavior modeling, semantic technologies, usage log analysis, linear temporal logic, model checking} 

\section{Introduction}
Understanding of user behavior in accessing Web resources is a powerful tool for Web sites providers to improve their applications, the design and content of the Web sites, to analyze users' navigation intentions and respectively improve search or build adaptive sites. Furthermore, it helps to  build recommendation systems that suggest users future actions based on their past behavior. 

The benefits of analyzing usage behavior analysis has been driving continuous research in the realm of Web Usage Mining, which aims at discovering navigation patterns from the logs of HTTP requests of pages and Web resources from the visitors of the sites. Due to the primarily syntactical nature of such resources, the representation of users' requests lacks a semantic formalization. This makes the comprehension of the mined patterns difficult. 

A useful step towards a better interpretation and analysis of the usage patterns is to formalize the semantics of Web resources and user browsing behavior. We focus on this problem and present in this paper an approach for the semantic formalization of user Web browsing activity. It lays the basis for effective techniques of querying expressive usage patterns, as well as more intelligent mining and recommendation methods.

In our approach, we map the records of usage logs issued by human visitors to meaningful events of the application domain where they were triggered. Therefore, instead of a syntactic representation, we now provide a semantic, formal description of each log by mapping it to concepts of a vocabulary of the domain knowledge. Instead of using flat taxonomies to represent such vocabulary, we address the use of ontologies to structure domain concepts and relations, ensuring a rich semantic model of the Web site content.

Nowadays there is an increasing number of semantically-enabled Web sites, which provide an ontology for the representation of their content and belonging Web resources. In this work, we assume that the domain ontology is already provided and, therefore, focus for now on the formalization of usage logs of these semantically-enabled Web sites. As such, we show the applicability of our approach using the USEWOD datasets \cite{usewod} of logs from two Linked Open Data servers: DBpedia~\footnote{http://dbpedia.org} and Semantic Web Dog Food (SWDF)~\footnote{http://data.semanticweb.org}.

The goal of our approach is the formalization of browsing behavior not only restricted at a single Web site, but rather at multiple sites. The absence of publicly available cross-site usage logs is another reason for choosing the USEWOD datasets. Despite the fact that these data are gathered independently from the two Web servers, we design our techniques with the aim of discovering patterns of cross-site behavior, investigating traces of users navigation from one Web domain (e.g. DBpedia) to another (e.g. SWDF, various search engines, etc.). 

Another crucial aspect in analyzing browsing behavior, is also its temporal dynamic related to the order of requests being issued. We present a query formulation and answering approach that is able to search for  expressive patterns in terms of temporal constraints. We apply a formalism that extends description logic with temporal constructs (such as what happens next, eventually, always) based on Linear Temporal Logic~\cite{TDL}.
We further show how to search for behavioral patterns upon the semantically formalized usage logs applying the query answering technique. The adaptation and application of this logic and techniques for the setting of Web usage analysis are novel. 

\begin{figure}[hb!]
	\centering
	\scalebox{0.57}{ \includegraphics{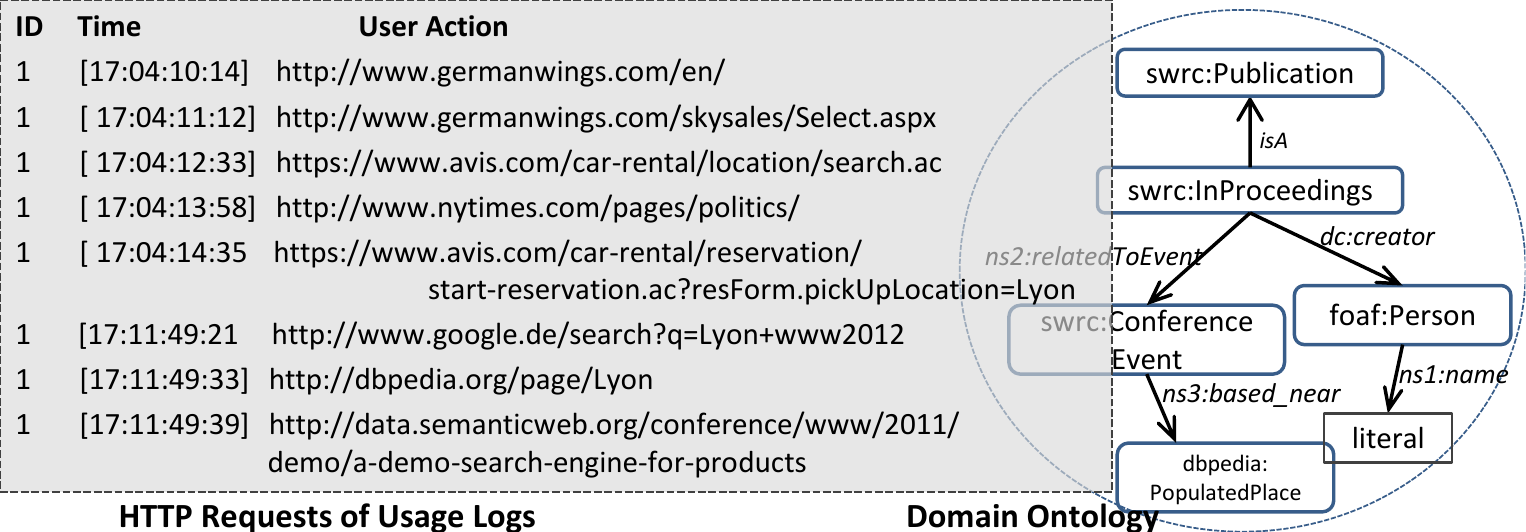}}
	\caption{User browsing logs and domain ontology}
	\label{fig:logs}
\end{figure}
\textbf{Example 1 (Usage Logs and Domain Knowledge)} \emph{In an example of usage logs (Fig.~\ref{fig:logs}), we show a user sessions in which the user starts looking for flights at \emph{germanwings.com}, afterwards searches for cars to rent at \emph{avis.com}, in between reads some news at \emph{nytimes.com}. She then continues further the reservation for car rental. Afterwards, the user performs a search at \emph{google.com}, followed by a visit of the \emph{Lyon} page at \emph{DBpedia}. At last, she browses a specific paper from \emph{WWW2011} conference at \emph{SWDF}.} 

In our approach, we map the entries of usage logs to concepts in the domain ontology (in this case illustrating that of SWDF), e.g. last log entry represents a \emph{swrc:InProceedings}~\footnote{http://ontoware.org/swrc/swrc\_v0.3.owl\#} of type \emph{swrc:Publication}. We can also extend the context for each log via the domain ontology, deriving in this case more information (e.g. author, conference, location, etc.) that we can use to query for more expressive usage patterns.

Leveraging usage data with semantics helps to find in this use case patterns, such as:
\begin{itemize} 
\item Find browsing sessions between March and April 2009 in which the users were browsing DBpedia or SWDF,
eventually performed an engine search, then immediately returned back to the previous site(i.e. goal is to detect possible bottlenecks in site design or search).
\item Find how often users have visited resources of English Musicians of Jazz genre (i.e. apply the extended context derived from the ontology in the background) .
\item Search for flight tickets at a given site, but then make reservation for car rental at another Web site (i.e. goal is to detect failing services from the perspective of a Web site provider).
\item Patterns of arranging a travel to a city, afterwards access a DBpedia page about that city, then look for a conference located in this city (goal is to predict what users might want to visit in the future based on (personal/collective) past behavior).
\item Find only semantically-relevant events in the usage patterns. e.g. filter out reading the news in the middle of a travel arrangement session, etc.
\end{itemize}

The contributions of this work are as follows:\\
- Formalization approach for cross-site user browsing behavior and specific techniques applied to formalization of usage logs from Linked Open Data servers (DBpedia and SWDF).

- A framework for querying expressive patterns of user browsing behavior with semantic- and temporal-based constraints. We introduce an approach for query formulation using temporalized description logics,  which allows us to reason about user behavior with temporal conditions. We also present the approach of answering such queries using model checking techniques(Sec. 3). To the best of our knowledge, these are techniques not previously applied in the setting of Web usage analysis.

- Experimental results that show the feasibility of our techniques and aslo serve as an exploratory analysis of the way users browse the Web of Data, in this case focusing our study on DBPedia and SWDF sites.

\begin{figure*}[ht!]
	\centering
	\scalebox{0.70}{ \includegraphics{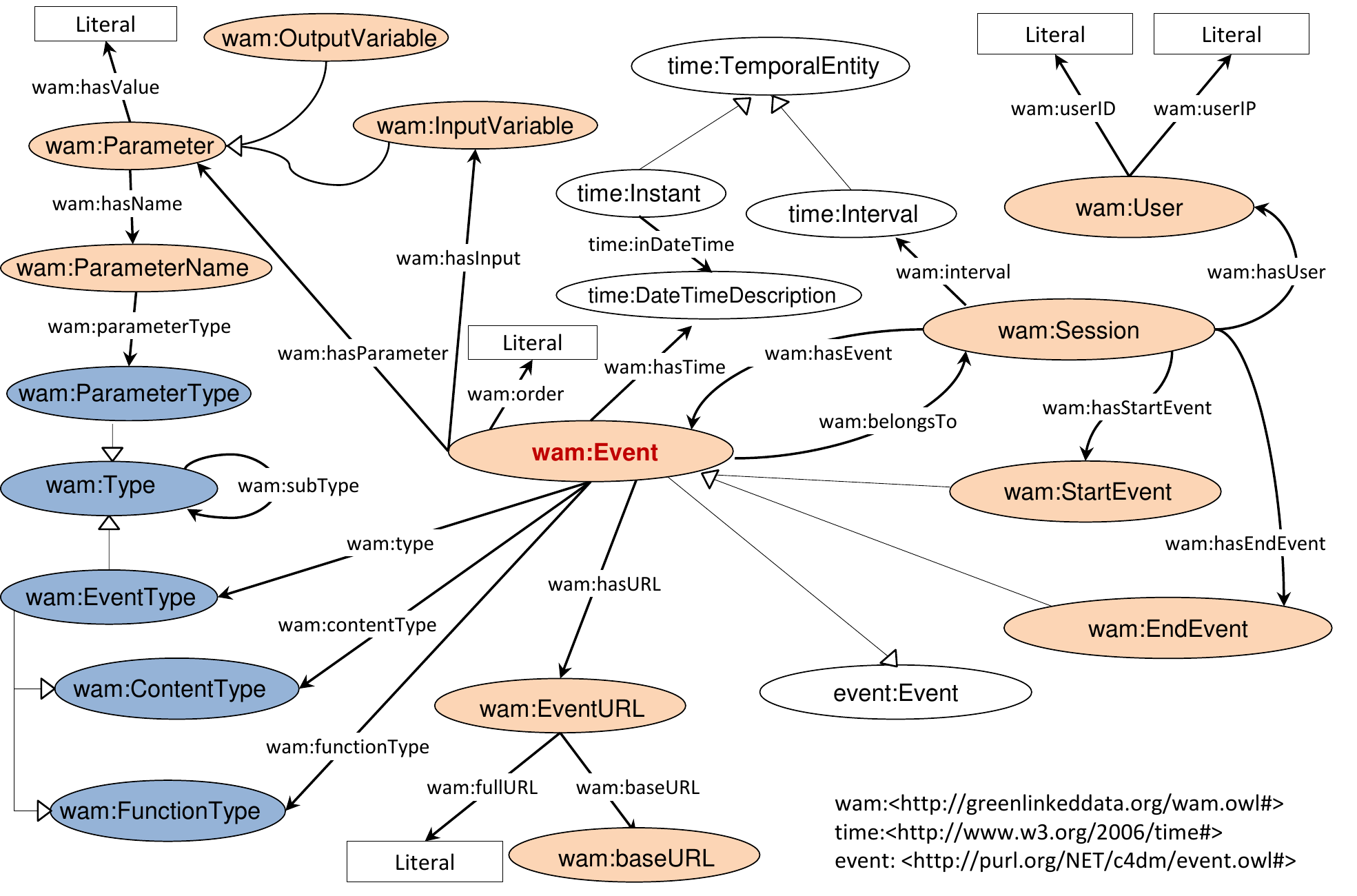}}
	\caption{Web Browsing Activity Model - WAM}
	\label{fig:WAMontology}
\end{figure*}

\section{Semantic Formalization of User Browsing Behavior}
When browsing or searching the Web, users interact with Web resources via browsers interface (e.g. clicking links,
submitting HTML forms, etc.). We use the term \textit{browsing event} to describe the basic component of user behavior in performing activities (actions) with the Web browser directly.
 
\begin{definition} {\emph{\textbf{(Event)}}} We define a browsing event as a tuple $e$ = $(l, \mathcal{T}, P, t)$, where $l$ is the full URL invoked, $\mathcal{T}$ is a set of event types for which this event qualifies, $P$=$\{p_1,...,p_m\}$ is a set of parameters and $t$ is the \emph{occurrence time}.  For simplicity, we denote event time by $e_i.t$ and set of event types by $e_i.\mathcal{T}$.
\label{def:event}	
\end{definition}
\vspace{-0.4cm}

Each event resulting from the interaction of a user with a specific Web page serves a particular function (searching within a portal, searching in a search engine, browse information, booking, login, etc.) related to some content (e.g. flight reservation, car rental, organization, person, hotel, etc.). 

\begin{definition} {\emph{\textbf{(Event Type)}}} An event $e$ can be mapped to two types denoted by the set $\mathcal{T}$ = $\{\mathcal{T}_c, \mathcal{T}_f\}$, where $\mathcal{T}_c$ is the type of content to which this event relates, $\mathcal{T}_f$ is the type of function this event $e$ serves.
\label{def:eventtype}	
\end{definition}
\vspace{-0.4cm}

Referring to the example in Fig.~\ref{fig:logs}, each log entry is regarded as a browsing event, e.g. $e_7$ = \emph{(http://dbpedia.org/page/Lyon}, $\mathcal{T}$, $P$, \emph{17:11:49:33)}, where content type $\mathcal{T}_c$=\emph{\{InProceedings\}} and event function type $\mathcal{T}_f$ = \emph{\{Informative\}}. Other examples of content type for a browsing event include person, organization, real estate, education, stock exchange, etc. Whereas, examples of function type could be login, search, checkout, reserve, etc.

We have extended the definition of a browsing event with parameters, which can be extracted based on the information contained in the URL $l$. We consider three main conceptual elements in a link: URL base, variable names, and values. Based on the typical convention of URL formation, we syntactically split the link into two basic parts: URL base, which defines
domain name, and the rest of the URL is used to extract input variables, which are modeled as event parameters.

\begin{definition} {\emph{\textbf{(Parameter)}}} An event parameter $p$, which can be further classified as input or output parameter~\footnote{In this work, we focus on the input parameters only}, is a pair $p$=$(v_{name}, v_{value})$ consisting of variable name and value. 
\label{def:parameter}	
\end{definition}
\vspace{-0.4cm}

In the example of Fig.~\ref{fig:logs}, the input parameter of event $e_5$ is $p$=\emph{(resForm.pickUpLocation, Lyon)}. Furthermore, events are grouped into sessions, which represent a period of sustained Web usage. The boundaries of a session are normally determined by temporal and behavioral factors (e.g. browsing intention). We follow previous research~\cite{Bucklin2003} in deploying an heuristic, which starts a new session after a user's idle period of \emph{30 minutes} between the browsing events.

\begin{definition} {\emph{\textbf{(Session)}}} We denote a user session as a tuple $S = \{s, T_s, T_e, U\}$, where $s$ = $\langle e_1,e_2,...,e_n\rangle$ is an ordered sequence of browsing events performed from user $U$, such that $e_i.t \leq e_{i+1}.t$ for all $i$, where $i$ denotes the event order in the sequence. Furthermore, $T_s$ is the starting time and $T_e$ the ending time of the session, such that $T_s \leq e_i.t \leq T_e$.
\label{def:session}	
\end{definition}
Ordering of the events in a session is used to later define the notion of abstract time, under which we formulate temporal constraints to query behavioral patterns.

\begin{figure*}[ht!]
	\centering
	\scalebox{0.65}{ \includegraphics{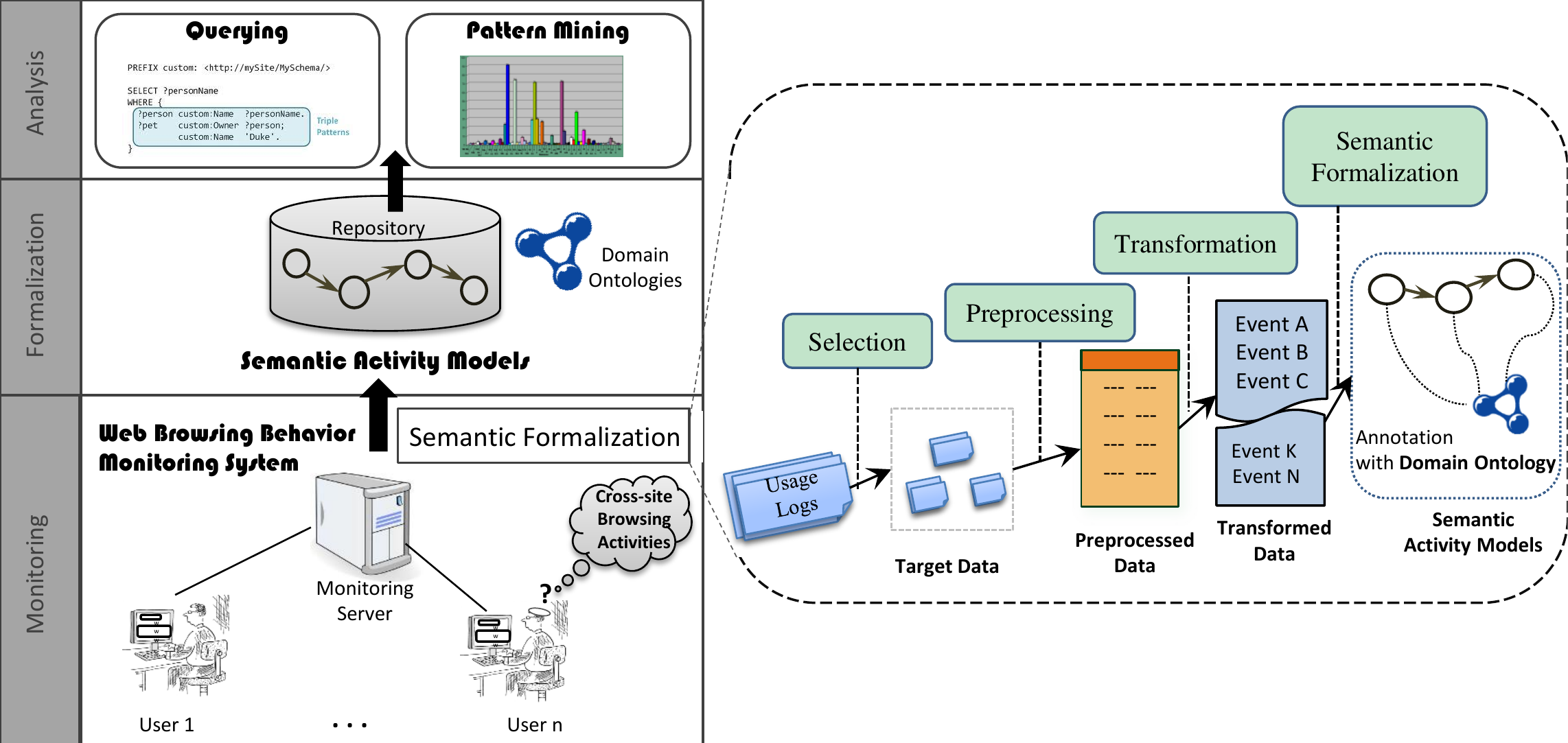}}
	\caption{Platform for semantic analysis of user browsing behavior}
	\label{fig:overall-approach}
\end{figure*}

\subsection{Web Browsing Activity Model}
For the realization of the concepts defined above, we present a Web Browsing Activity Model (WAM)~\footnote{http://greenlinkeddata.org/wam.owl}, which we formalize as an ontology illustrated in Fig.~\ref{fig:WAMontology}. Classes in WAM  are divided into three groups: Core classes, External classes, and Type classes.
External classes are basic concepts that we reuse from well-established ontologies. Each \texttt{wam:Event} is a subclass of the concept \texttt{event:Event} from the \textit{Event ontology}~\footnote{http://purl.org/NET/c4dm/event.owl\#}.
Each \texttt{wam:Session} has one \texttt{wam:StartEvent} and one \texttt{wam:EndEvent}, both of type \texttt{wam:Event}. Class \texttt{wam:User} is simply characterized by user IP address and ID, but the ontology allows flexible future extendability with user profiles or other attributes (e.g IP-based geographical location). 
To annotate of event timestamps and session interval, we reuse basic concepts from \textit{OWL Time ontology},~\footnote{http://www.w3.org/2006/time\#} which models knowledge about time such as temporal units, instants, etc.
The WAM ontology is expressed in OWL-2-DL with underlying $\mathcal{SROIQ}$ logic~\cite{HKS06-sroiq}. 

\textbf{Semantic Enrichment using Domain Knowledge}\\
Each user browsing activity recorded in logs is physically represented by a URL, but conceptually it comprises an event that
serves a particular function and relates to a specific content. We give meaning to each event issued as an HTTP request in the logs,
by mapping its respective URL to the WAM concepts according to two dimensions: content and function, respectively modeled as \texttt{wam:ContentType} and \texttt{wam:FunctionType}. In order to find the correct semantics of the event types, we refer to the knowledge of the domain (Web site) where this event occured. In this paper, we refer to the domain ontologies~\footnote{It is important to distinguish domain ontology from our WAM ontology, which we present for the formalization of usage logs.} and RDF representations of Web resources available for DBpedia and SWDF.
In the following section, we present the approach that we apply for the formalization of such browsing events.

\subsection{Formalization Approach}
Our overall approach (illustrated in Fig.~\ref{fig:overall-approach}) aims at monitoring user browsing behavior across multiple Web sites. Each log entry is a tuple $\mathcal{L} = \{UserID, URL, timestamp\}$ consisting of a user ID, referring momentarily to the encoded IP address of the user, URL of the accessed Web resource, and real time when this happened.

Firstly, these usage data logs are centrally stored in raw form as produced upon each user interaction. We regard each entry log
$\mathcal{L}$ as a browsing event, as defined in Def.~\ref{def:event}.  These events are grouped into sessions based on user IP address. The formalized browsing events constitute semantic-rich models of user browsing activity, which we store in a
\emph{repository}. This repository is then used as foundation for the semantic analysis of usage behavior, including query answering techniques, which we present in Section 3, and further potential semantic pattern mining mechanisms.

Below we present the formalization algorithm, which takes as input a set of raw logs, preprocesses these data following the engineering steps also shown in Fig.~\ref{fig:overall-approach}. 
\newcommand{\Od}{$\mathcal{O}_b$}
\newcommand{\Ow}{$\mathcal{O}_{wam}$}
\newcommand{\e}{$\texttt{e}$}
\newcommand{\p}{$\texttt{p}$}
\newcommand{\Pname}{$\texttt{P}$}

\newcommand{\INDSTATE}[1][1]{\STATE\hspace{#1\algorithmicindent}}

\begin{algorithm} [ht!]
\label{alg:formalize}
\centering
\caption{Main Algorithm for Formalization of Web Browsing Events}
\begin{algorithmic}
\REQUIRE Web server log entries $\mathcal{L}$=${L_1,L_2,...}$ where $L_i=(IP, t, URL, Referring URL)$
\ENSURE \emph{\tdl} Update knowledge base with new ABox assertions $\mathcal{A}$ for sessions and events 
\STATE \textbf{Step 1. Clean logs $\mathcal{L}$ from records of bots, in order to filter only human-generated requests}
\STATE \textbf{Step 2. Group logs in sessions based on user IP address}
\STATE \textbf{Step 3. Filter logs }
\INDSTATE Introduce a new event for each session of $L_i$ that contains a referring URL from another Web domain
\STATE \textbf{Step 4. Formalize browsing sessions based on the WAM ontology}
\INDSTATE For each log entry $L_i$ create a session of browsing events $s=\langle e_1,e_2,...,e_n\rangle$

\FORALL{$e_i \in s$, where $e_i=(U, l, t)$}
\INDSTATE $b=extractBaseUrl(l);$
\INDSTATE Assertions $\alpha_{e}$ = $\{Event($\e$_i)$, $fullURL($\e$_i,e_i.l)$, 
\INDSTATE  $baseURL($\e$_i,e_i.b)$, $ time($\e$_i,e_i.t), order($\e$_i,j$++$)\}$ 
\INDSTATE Find event content types $\alpha_{t}$=\emph{findContentTypes(l, b)}
\INDSTATE Find function types and parameters, and create assertions $\alpha_{f}$, $\alpha_{p}$, accordingly.
\ENDFOR
\INDSTATE $\mathcal{A} = \alpha_{e} \cup \alpha_{t} \cup \alpha_{f} \cup \alpha_{p} $
\STATE \textbf{Step 4.1 Serialize $\mathcal{A}$ for sessions and events in RDF/XML}
\STATE \textbf{Step 5. Update knowledge base (Sesame repository of semantic logs) }
\end{algorithmic}
\end{algorithm}

\begin{algorithm} [ht!]
\label{alg:formalize}
\centering
\caption{Semantic Enrichment of Browsing Events \emph{findContentTypes(l, b)}}
\begin{algorithmic}[3]
\REQUIRE Ordered sequence of events $s=\langle e_1,e_2,...,e_n\rangle$
\ENSURE ABox assertions $\alpha_{t}$ related to content types 
\STATE \textbf{Given URL $l$ and its Web domain $b$ }
\STATE \Od $= getDomainOnt(b);$
\STATE Resource $R_l=findResource(l,$\Od$);$ 
\STATE We configure for each domain (DBpedia, SWDF) a template of mappings from Web resources to their RDF/XML representation e.g. for SWDF add '/rdf' to a link that is a URI, or replace 'html' with 'rdf' for a link that is an HTML page.
\STATE  
\STATE Find classes $\mathcal{T}_c=classMembership(R_l);$ We read the RDF representation of the resource and find classes (using rdf:type) to which it belongs. 
	\FORALL{$T \in \mathcal{T}_c$}
		\STATE	$\alpha_{t} = \{ContentType(T), contentType($\e$_i,  T)\}$
	\ENDFOR
\end{algorithmic}
\end{algorithm}
\vspace{-0.4cm}
The algorithm yields as result a set of the ABox assertions related to each event, based on the TBox concepts shown in the WAM ontology. We focus here on the approach used to annotate the content type of each event, using the RDF/XML representation of the Web resource requested (Alg. 2, line 3). In the end, the generated DL axioms and assertions are added in the knowledge base, which is a triple-store repository of these semantic usage logs.

\section{Querying User Browsing Patterns}
It is important to reason about the browsing behavior of users not only using conditions related to the enriched semantics, but also addressing temporal constraints regarding the dynamics of such behavior. To achieve this, we allow queries upon our knowledge base to be formulated involving temporal operators such as \textsf{eventually, until, always}. 

This motivates us to address temporal logics capable of ontological reasoning. To support such temporal reasoning, we follow an approach similar to the one in ~\cite{ALC-LTL}, which presents the temporalized description logic $\mathcal{ALC}$-LTL as an extension of $\mathcal{ALC}$ with Linear Temporal Logic (LTL). 
Instead of $\mathcal{ALC}$, we apply in our approach $\mathcal{SROIQ}$~(as in \cite{HKS06-sroiq}) the DL underlying OWL-2-DL, which used to express our WAM ontology. We refer to this formalism as \emph{\tdl} and use it for formulation of queries upon the formalized browsing events and sessions.


\subsection{Query Formulation}
\label{sec:queryformulation}
In our querying approach, we consider the notions of real time and abstract time. As mentioned in Def.~\ref{def:session} of a session, the ordering of events within the session is used as basis for abstract time to formulate temporal constraints. As an example, let's consider the following query:

Query $q_1$: Find sessions where user starts browsing \emph{homepage} of Web site $w_a$, then \emph{eventually} performs \emph{search } in an engine (site), afterwards returns back to site \emph{$w_a$}.

Answering $q_1$ requires finding a session with a start time $T_s$ and end time $T_e$, and also the set of events $e_i \in S$ s.t. $T_s \leq e_i.t\leq T_e$. Within this set of events, we filter only those events having the required URLs and order. The former part of the query deals with the real time, whereas the latter temporal part deals with the abstract time (ordering of events in a timeline). 

This query can be expressed as a ~\tdl ~formula, whose notion we define inductively as follows:
\vspace{-0.5cm}
\begin{itemize}
\item every ABox assertion is a \tdl ~formula. For example,  $\mathtt{Event}(e)$ and $\mathtt{hasURL}(a,b)$  are \tdl ~formulas.
\item if $\varphi$ and $\psi$ are \tdl ~formulas, then so are $\varphi\wedge\psi$, $\varphi\vee\psi$, $\neg\varphi$, $\varphi\mathtt{U}\psi$, and $\mathtt{X}\varphi$ ( $\mathtt{U}$ being the $Until$ operator and $\mathtt{X}$ the $Next$ operator).
\end{itemize}
\vspace{-0.4cm}
As an example, the \tdl\ ~formula  
\begin{align*}
\mathtt{Event}(e_1)\wedge\exists~\mathtt{contentType.Publication}(e_1))\wedge\\
\mathtt{X}(\mathtt{Event}(e_2) \wedge \exists \mathtt{functionType.SearchEngine}(e_2))
\end{align*}
describes an event $e_1$ with content type \emph{Publication} and followed by an event $e_2$ of event function type \emph{search engine}.

For query answering, we need to check if a temporal pattern is satisfied within a session. By Definition~\ref{def:session}, every session $S=(s, T_s, T_e, U)$ contains a sequence of events $s=\langle e_0,\dots,e_n\rangle $ with $T_s\leq e.t\leq T_e$. Since the temporal operators refer to the ordering of the events in this sequence, we consider this order to be the timeline of the abstract time. 

As mentioned above, we are interested in queries that not only requires checking the satisfaction of temporal patterns in a session, but also the satisfaction of certain conditions on the session itself. Similar to the description of an event, a session is also described with a set of ABox assertions. 
We define an \emph{atom over a given variable $x$} as an assertion parametrized with variable $x$, i.e., atoms are of the form $C(x)$, $R(x, a)$ or $R(a, x)$ for a concept $C$, a property $R$ and a constant $a$. We define a query as follows:
\begin{definition}\label{def:query}
Let $\omega$ and $\varphi$ be conjunctions of atoms over a some variable $x$. A query $Q$ over $x$ is an expression of the following form
\[Q(x) \leftarrow \omega,\varphi\hspace{1.5cm}(1) \]
s.t. $\omega$ denotes the set of assertions related to the session, and $\varphi$ is a \tdl~formula representing the temporal pattern related to the events within the session. 

Given a set of sessions $\{S_1,\dots,S_l\}$ along with an ontology $\mathcal{O}$, the answer  $\mathtt{Ans}(Q)$ to a query $Q$ of form $(1)$ over  variable x is a subset of $\{S_1,\dots,S_l\}$ such that for each $S\in\mathtt{Ans}(Q)$ we have $\mathcal{S}(A)\models\omega[x/S]$, and  $\mathcal{S}(A),0\models\varphi[x/S]$, \\
where $\omega[x/S]$ ($\varphi[x/S]$) represents the conjunction of assertions obtained from $\omega$($x$)  by replacing the variable $x$ with $S$.
\end{definition}
\vspace{-0.4cm}

We separate conditions (to be satisfied) on a session from the temporal patterns (to be verified) within the session in order to make the formulation of the query more understandable. Below we give examples of three queries and their formulation \emph{\tdl}. \\
Query $q_1$: Find sessions where user starts browsing  \emph{homepage} of Web site $w_a$, then \emph{eventually} performs \emph{search } in an engine (site), aftewards returns back to site \emph{$w_a$}.
\vspace{-0.2cm}
\begin{align*}  
	q_1(s) \leftarrow (& \mathtt{Session}(s) \wedge \mathtt{hasEvent}(s, e_1) \\
							& \wedge\mathtt{hasEvent}(s, e_n) \wedge \mathtt{hasEvent}(s, e_m)), \\
						  &( \mathtt{Event}(e_1)  ~\wedge \mathtt{baseURL}(e_1,w_a) \wedge \\
							&\exists \mathtt{functionType.Homepage}(e_1)) ~\wedge\\
							&\Diamond ( (\mathtt{Event}(e_m)  \wedge \exists \mathtt{functionType.EngineSearch}(e_m)) ~\wedge\\
							&\mathtt{X}(\mathtt{Event}(e_n)  \wedge\mathtt{baseURL}(e_n, w_a)))\\
\end{align*}
\vspace{-1.1cm}

Query $q_2$: \emph{Find sessions where user browses site related to \emph{music}, then eventually performs a search in site \emph{google.com} or site \emph{yahoo.com}, then immediately comes back to a \emph{page of a music group} in site $w_b$ }
\begin{align*}  
	q_2(s) \leftarrow (& \mathtt{Session}(s) \wedge \mathtt{hasEvent}(s, e_1) \\
							& \wedge \mathtt{hasEvent}(s, e_m) ~\wedge \mathtt{hasEvent}(s, e_n)),\\
	 					  & (\mathtt{Event}(e_1) \wedge \exists~\mathtt{contentType.Music}(e_1)) ~\wedge \\
							& \Diamond ( (\mathtt{Event}(e_m)   \exists \mathtt{functionType.EngineSearch}(e_m) ) ~\wedge \\
							& \wedge (\mathtt{baseURL}(e_m,``http://www.google.com``) \\
							& \vee  \mathtt{baseURL}(e_m,``http://www.yahoo.com``) )~\wedge\\
							&  \mathtt{X}(\mathtt{Event}(e_n)  ~\wedge ~\mathtt{baseURL}(e_n, ``w_b``) \wedge\\
							&\exists~\mathtt{contentType.MusicGroup}(e_n) )) \\
\end{align*}
\vspace{-1.1cm}

Query $q_3$: \emph{Find sessions where user is browsing all the time sites $w_a$ or $w_b$}
\begin{align*}  
	q_3(s) \leftarrow & ( \mathtt{Session}(s) ~\wedge \mathtt{Event}(e) ~\wedge \mathtt{hasEvent}(e) ),  \\
							& \square (\mathtt{baseURL}(e,``w_a``) ~\vee \mathtt{baseURL}(e,``w_b``))
\end{align*}

In the following section, we introduce the approach we apply for answering such queries.

\subsection{Query Answering Approach}


\renewcommand{\vec}[1]						{\ensuremath{\mathbf{ #1 }}}
\newcommand{\muAlways}[1]                   {\ensuremath{ \mathbf{ always } \ #1 }}
\newcommand{\muEventually}[1]               {\ensuremath{ \mathbf{ eventually } \ #1 }}
\newcommand{\muUntil}[2]                    {\ensuremath{ #1 \ \mathbf{ until } \ #2 }}
\newcommand{\muExistsAction}[2]             {\ensuremath{ \langle #1 \rangle #2 }}
\newcommand{\muForAllAction}[2]             {\ensuremath{ [ #1 ] #2 }}
\newcommand{\muMinFixPoint}[2]              {\ensuremath{ \mu #1 . #2 ( #1 ) }}
\newcommand{\muMaxFixPoint}[2]              {\ensuremath{ \nu #1 . #2 ( #1 ) }}
\newcommand{\muTrue}                        {\ensuremath{ \top }}
\newcommand{\muFalse}                       {\ensuremath{ \bot }}

We present a query answering approach based on a matchmaking technique, which allows to automatically retrieve sessions of user browsing events that satisfy a set of semantic and temporal conditions. These conditions are formulated in a query $q$, which as defined in Def.~\ref{def:query} is an expression of the form  $q(x) \leftarrow \omega,\varphi$, s.t. $\omega$ consists of constraints related to the session itself, and $\varphi$ is a \tdl~formula representing the temporal pattern related to the events within the session. The query is posed upon the set of sessions $\mathcal{S}$, which are formalized with the approach described in Section 2 and that we have stored in a central repository.  In order to find the answers to the query, we apply a matchmaking mechanism that proceeds as follows:

\paragraph*{Step 1 - Checking Satisfiability of Session Constraints}
This steps deals with expression $\omega$ of the query $q$, which defines constraints related solely to the attributes of the session.  As explained earlier, $\omega$ consists of a conjunction of parametrized ABox assertions $\alpha_s$. In this step, we check for each of the sessions in the repository if it satisfies the assertions, using HermiT reasoner~\footnote{http://hermit-reasoner.com/}.


For example, in this step we would retrieve sessions with starting time $Ts$ in July and $userIP$ of a particular value $IP$.

\paragraph*{Step 2 - Checking Satisfiability of Temporal Constraints}

The remaining part of the query $Q$ consists of the \tdl~formula $\varphi$, which defines the temporal constraints over the events contained in the sessions that we retrieved in Step 1. Therefore, only the resulting set of sessions from step 1 (denoted as $S^1$) are now considered in Step 2.
%

In order to check the satisfaction of temporal constraints, we iterate over the sessions in $S^1$ and (a) build a finite state automaton (FSA) for each $S_i \in S^1$, afterwards (b) iterate over the states of the automaton in order to determine whether a condition holds in the respective state. In the sequel, we provide more details about this verification technique, which applies a model checking mechanism based on the modal $\mu$-calculus~\cite{kozen83}. The goal is to check if the formula $q$ holds in the session $S_i \in S^1$ represented as a sequence of events. 

\textbf{Step 2(a) - Construction of the FSA} 

The semantics of a \tdl{} formula $\varphi$ with temporal constraints is defined over a finite-state transition system that represents the sequence of events within a session $S_i$. 
Let $F_{S_i} = (\mathcal{P},p_0,\pi )$ denote an FSA representing the sequence of events in session $S_i$. The automaton is described by a final set $\mathcal{P}$ of states, a unique start state $p_0 \in \mathcal{P}$, and a transition function $\pi : \mathcal{P} \rightarrow \mathcal{P}$.
Each state corresponds to a separate knowledge base (set of DL axioms of OWL ontology) that describes one event of the sequence, while the event sequence order is preserved by the transition function. 

We construct a $F_{S_i}$ from each sequence of events in the session: the start state $p_0 \in \mathcal{P}$ of $F_{S_i}$ is generated by adding the axioms of WAM ontology $O_A$ that describes the domain terminology used for events (TBox) as well as ABox assertions related to the session itself and its first event (with order 1). 

The subsequent states and transitions are generated according to the event sequence in $S_i$. For a sequence $\langle e_i, e_{i+1} \rangle$ of two subsequent events $e_i$ and $e_{i+1}$, a transition $\pi = (e_i, e_{i+1})$ is added from the state $p_i$ to a new state $p_{i+1}$. State $p_{i+1}$ is created by adding the description of the event $e_{i+1}$ and the static domain knowledge from $O_A$.

\textbf{Step 2(b) - Verification of \tdl ~formula in $q$}

For a given automaton $F_{S_i}$ and a given the \tdl ~formula  ~$\varphi$, the matchmaker findes the subset of states of $F_{S_i}$ in which the formula is satisfied. In case ~$\varphi$ is defined as a composite formula, then it is broken down into a set of atomic formulas $\phi_i$. The final result is aggregated from the results of the atomic formulas recursively according to the semantics of the query formalism. We proceed as follows for each of the atomic formulas:\\ 
If $\phi = \muTrue$ (true), then all the states of $F_S$ are returned.\\ 
If $\phi = \muFalse$ (false), then an empty set is returned. \\
If $\phi = \Psi$ (proposition $\Psi$ over a desired event description) then according to the semantics of the query formalism, we need to find those states in which the proposition $\Psi$ holds, i.e., the desired event occurred. 

Iterating over all the states of the automaton $F_S$, we add a state $p$ in the resulting set, if the proposition $\Psi$ holds in state $p$. Since a state is an OWL ontology and a proposition is a data query~\footnote{Currently we only support conjunctive queries on ontologies with our own implementation of query evaluation based on HermiT reasoner.}, we execute the data query on the OWL ontology. If the result set of the data query is non-empty, then the proposition holds in the state, otherwise not. 
A detailed explanation of the matchmaking algorithm is presented in~\cite{agarwal-ws2008,agarwal-WI2007}.
The end result of this verification technique is a set of sessions (URIs), in which the sequence of events satisifies the \tdl{} formula part of the query $q$.
\newcommand{\cell}[1]{\;\; #1\;\;}

\section{Implementation and Evaluation}
We provide a Java SE implementation of the introduced formalization approach, deploying the steps of processing usage logs, cleaning, and formalization with WAM ontology (whose consistency is checked with Pellet 1.5.2 reasoner). We have further implemented the step of semantic enrichment of events with predefined ontologies of Web domains, which we read and query using Jena Framework~\footnote{http://incubator.apache.org/jena/}. 
In order to show the feasibility of our formalization approach, we have performed experiments in which we have processed and semantically formalized logs from the datasets as featured in Table~\ref{tab:eva-datasets}. The formalized sessions and events are serialized in RDF representations, which are afterwards imported via OpenRDF Sesame Core 2.6.0 API~\footnote{http://www.openrdf.org/doc/sesame2/api/} into a repository of a Sesame Framework~\footnote{http://www.openrdf.org/}.  This repository is available online~\footnote{http://46.4.66.131:8080/openrdf-workbench/repositories/wam/query} for the interested reader to test the semantic models. Because of privacy restrictions of the datasets, we have imported only a small portion of the formalized logs in this repository. 





\vspace{-0.4cm}
\begin{table}[tb]
\caption{Results of the Formalization Approach\label{tab:eva-datasets}}
~\hfill\begin{tabular}{|p{2.5cm}|p{1.3cm}|p{1.5cm}|} \hline
    &  SWDF 2009 & DBPedia 3-3\\
\hline\hline
\#sessions &   $2831$  &  $31893$ \\
\hline
avg.\#sessions/day &   $235.92$   &  $2899$ \\
\hline
\#triples &   $277788$  &  $>3$million  \\
\hline
\hline
Monitoring Period & 01.Jul.09-12.Jul.09 & 01.Jul.09-12.Jul.09\\
\hline
\end{tabular}\hfill~
\end{table}

Regarding the content of the datasets, we observed that SWDF has between 161 and 309 user sessions daily, with an average of 235.9 sessions/day. The majority of the daily sessions (57.2\%) start at search engines and then continue at SWDF, while 33.5\% start with the user directly accessing SWDF. From the daily sessions initiated in search engines, an average of 97\% start at Google (.com,.de, etc.) and only 2.7\% at Bing. There are very few sessions from human users containing SPARQL queries (in average 1.46\% daily sessions). The majority of sessions containing SPARQL queries belong to machines/bots, which we have filtered out of our usage data formalization approach. 

We have also performed experiments to test the feasibility of the query answering approach. We formulated and executed a set of \tdl ~queries on the repository of formalized events. These queries are illustrated in Table~\ref{tab:eva-queries}.  We used an IBM Thinkpad T60 dual core, with 2 GHz per core, Windows 7 (32-bit) as the operating system, and a total of 2 GB memory.
\vspace{-0.4cm}
\begin{table}[ht]
\caption{Queries used for experiments\label{tab:eva-queries}} 
{\small
 \begin{tabular}{|l|p{6cm}|c|c|}
  \hline
$\cell{\mbox{Query} }$  & Description   \\
\hline\hline  
    $\cell{Q_1}$    &  find all sessions starting with a search in a search engine, then followed by a browsing event in DBpedia  \\ \hline
    $\cell{Q_2}$    &  find all sessions starting with a browsing event in SWDF, followed by a search in google.com \\ \hline
    $\cell{Q_2}$    &  find all sessions where users have visited pages of Tango Musicians in DBpedia  \\ \hline
    $\cell{Q_4}$    &  find all sessions where users have browsed papers of the conference WWW2009 in SWDF, and then the page of Madrid in DBpedia\\  \hline
    $\cell{Q_5}$    &  find all sessions where the users have eventually visited English Artists in DBPedia	\\  \hline
\end{tabular}
}
\end{table}

In order to check the performance and scalability of the query answering mechanism, we have performed tests with different number of sessions, which is a portion of those saved in the repository.
We observed that the answering time varies slightly for the queries ($\sim$ 0.15 seconds), starting with 0.5 seconds for query execution in 500 sessions. For the largest number of sessions (up to 1000) upon which we performed tests, the answering time is still feasible and always below 1.4 seconds. 

\begin{figure}[ht]
	\centering
	\scalebox{0.95}{ \includegraphics{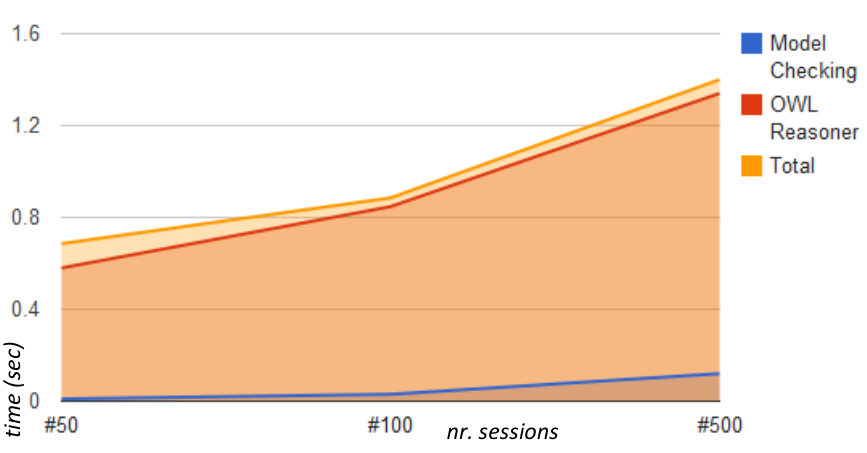}}
	\caption{Query answering time (composed of OWL reasoning time and model checking time) for one query and different nr. of session}
	\label{fig:eval_diagram2}
\end{figure}

In the diagram of Figure ~\ref{fig:eval_diagram2}, we show query answering time, reporting separately the time it takes for the OWL reasoning and for the model checking when testing it for one query ($Q_1$) and different number of sessions. We observe that model checking time is minimal, and reasoning takes nearly 94\% of the overall answering time.  

Overall, we observe that the query answering mechanism performs within feasible times (taking into consideration that we have not applied indexing, optimization techniques, parallelization, etc. yet).

\section{Related Work}
The works related to ours may be grouped as follows:

\textbf{Modeling cross-site browsing behavior.} 
Interest to characterize online behavior has started much earlier with works such as those of Catledge \emph{et al.} ~\cite{Catledge95}, Montgomery \emph{et al.}~\cite{Montgomery2001} that try to identify browsing strategies and patterns in the web. Browsing activity has been studied and modeled, e.g. Bucklin \emph{et al.}~\cite{Bucklin2003} and others, usually exploiting server-side logs of visitors in a specific website. Regarding the modeling of browsing behavior at multiple websites, Downey \emph{et al.} ~\cite{Downey07} propose a state machine representation for describing search activities. Park and Fader ~\cite{ParkFader2004} present a statistical model of browsing behavior in two Web sites. 


\textbf{Ontologies in Web usage mining.} 
There is an extensive body of work dealing with usage log analysis and mining, but we focus on the combination of these techniques with semantic technologies, which start with contributions of works such as those of  Stumme \emph{et al.}~\cite{Stumme02} and Oberle \emph{et al.}~\cite{Oberle2003}. In this field, research has been mostly focused on search query logs or user profiling.  Recent approaches, which use semantics for extracting behavior patterns from web navigation logs, are presented by Yilmaz \emph{et al.}~\cite{Yilmaz2010} and Mabroukeh \emph{et al.} ~\cite{Mabroukeh2009}.  While Yilmaz combines ontology and sequence information for sequence clustering, and Mabroukeh investigates sequential pattern mining and next step prediction.  Vanzin \emph{et al.} ~\cite{Vanzin2005} present ontology-based filtering mechanisms for retrieval of Web usage patterns, and more recently Mehdi \emph{et al.} ~\cite{Mehdi2010} tackle the problem of mining meaniningful usage patterns and clarify the impact of ontologies to solve this problem. These works are restricted to only one domain and not cross-site browsing behavior. 

\section{Conclusions and Future Work}
In this paper, we present an approach for the formalization of user Web browsing behavior across multiple sites, where we map
usage logs to comprehensible events from the application domain.
We have implemented this formalization approach and performed experiments with real-world datasets of usage logs from two Linked Open Data Servers (DBpedia and Semantic Web Dog Food).

Additional dynamic aspects of user browsing behavior can be discovered if reasoning not only with semantic constraints, but also
with temporal conditions is enabled. For this purpose, we introduce
an approach to formulate queries using a temporalized description
logic called \tdl{}, which combines $\mathcal{SROIQ}$~ with Lineal Temporal Logic (LTL). Alongside the formalism, we present a query answering mechanism, which is based on a model checking technique . This allows to automatically retrieve sessions of user browsing events that satisfy a set of semantic and temporal conditions. We show the feasibility of our approach through evaluations with usage logs from DBpedia and SWDF.

We plan to further investigate techniques related to the acquisition of the domain ontology when this is not provided, since it is a crucial component of the semantic enrichment of usage data with
concepts from the application domain. We will also work on pattern mining and prediction techniques, which will use as basis the semantically formalized events generated from this work. 

\section{Acknowledgment}
This work is partly funded by the EU-FP7 Project X-LIKE (Ref. 288342).

%
\bibliographystyle{abbrv}
\bibliography{sigproc}  
%
%
\end{document}